\documentclass[12pt]{article}

\usepackage{sbc-template}
\usepackage{graphicx,url}
\usepackage[utf8]{inputenc}

\usepackage{cite}
\usepackage{amsmath,amssymb,amsfonts}
\usepackage{algorithmic}
\usepackage{graphicx}
\usepackage{textcomp}
\usepackage{xcolor}
\usepackage{multirow}
\usepackage{bm}
\usepackage{amssymb}
\usepackage{subfigure}
\usepackage{fancyhdr}
\usepackage{url}
\usepackage[table]{xcolor}
\definecolor{cinza}{gray}{0.9}

\definecolor{DarkGreen}{RGB}{0,125,0}
\definecolor{DarkRed}{RGB}{125,0,0} 
\definecolor{DarkBlue}{RGB}{0,0,180}

\fancypagestyle{firstpage}{
  \fancyhf{}
  \fancyhead[C]{\textit{Preprint of paper accepted at SBBD 2026 -- Brazilian Symposium on Databases}}
  
}

\title{Gaussian Rank-Based Neighborhood Degree for Graph Neural Networks in Image Classification}

\author{Rafael Mendonça Duarte, Jean Roberto Ponciano, Lucas Pascotti Valem}
\address{Institute of Mathematics and Computer Science (ICMC) \\
University of São Paulo (USP) \\
São Carlos -- SP -- Brazil
\email{rmduarte@usp.br, jeanponciano@icmc.usp.br, lucas@icmc.usp.br}
}

\begin{document} 

\maketitle
\thispagestyle{firstpage}

\begin{abstract}

The exponential growth of data has intensified the gap between the availability of unlabeled data and the high cost of manual annotation. Graph Neural Networks (GNNs) have emerged as a promising solution, as they exploit relational structures and learn from both labeled and unlabeled data, performing semi-supervised learning.
A crucial component of many of these models is degree-based normalization, which influences message propagation but typically assumes uniform importance among neighboring nodes.
In image classification, graphs are usually constructed from feature similarity, where treating all neighbors equally may overlook important variations in relevance.
Motivated by this gap, we propose GRaNDe (Gaussian Rank-based Neighborhood Degree). This novel degree measure integrates neighborhood ranking with Gaussian distance weighting to better capture node importance.
Experiments on five public image classification datasets show consistent accuracy improvements and competitive or superior results compared to state-of-the-art methods.
The code will be made available upon paper acceptance.
\end{abstract}

\section{Introduction}

Advances in sensing, data acquisition, and storage technologies have dramatically expanded the volume of data available, and this growth continues to accelerate over time~\cite{AMINI2025128904}. While the amount of available data has been increasing exponentially, the process of labeling this data has not followed the same trend. Data annotation is both time-consuming and costly, which limits the availability of labeled samples in many real-world applications~\cite{AMINI2025128904}.
As a result, semi-supervised learning strategies have gained increasing attention, as they aim to exploit labeled and unlabeled data during the learning process~\cite{sbbdVeryFast}.

In this context, graph-based representations offer a natural way to encode sample relationships by representing samples as vertices and their relationships as edges, enabling the modeling of complex dependencies and structural patterns beyond pairwise feature comparisons. 
Information can then be iteratively aggregated from local neighborhoods, enabling the capture of both local and global patterns in the data.
Such representations have been exploited in different scenarios. For example, preference graphs have been employed for diversity-aware ranking fusion in image retrieval~\cite{sbbdGraphFusionRank}, and heterogeneous graph embeddings combining local attributes and neighborhood information have been proposed to enhance node representations~\cite{Angonese_Galante_2026}.

Among graph learning approaches, Graph Neural Networks (GNNs) have emerged as particularly promising, as they are able to effectively exploit the relational structure of data and propagate label information from a small set of labeled instances to a large number of unlabeled ones~\cite{JIANG2024112271}. To fully exploit these properties, GNNs provide a principled framework that leverages graph-structured data by iteratively aggregating neighborhood information, enabling the model to capture both local and global structural patterns in a unified and scalable manner.

Originally, GNNs were developed for data that is naturally represented as graphs, such as citation networks, social networks, biological interaction graphs, and molecular structures, where entities and their relationships are explicitly defined~\cite{JIANG2024112271}. In recent years, their applicability has increasingly expanded to multimedia data, particularly for visual data such as images, which can be transformed into graphs by modeling patches or feature descriptors as vertices and defining edges based on similarity, spatial proximity, or semantic relations. This representation allows GNNs to incorporate contextual structure beyond pairwise feature comparisons, benefiting tasks such as image classification, especially when labeled data is scarce.

Most GNNs, especially Graph Convolutional Networks (GCNs), are sensitive to the underlying graph topology~\cite{Khemani2024}. The way information is propagated across vertices plays a central role in the stability and effectiveness of these models~\cite{ICLR2025_a3017a8d}. In this context, degree-based normalization is a fundamental component of many GNN architectures, such as Approximate Personalized Propagation of Neural Predictions (APPNP)~\cite{paperGCN-APPNP2019} and Simple Graph Convolution (SGC)~\cite{paperGCN-SGC2019}, as it regulates how much influence each node and its neighbors exert during the message-passing process.

Given the importance of degree-based normalization in controlling message propagation, several studies have explored alternative approaches of node importance to replace or complement the standard degree measure, including eigenvector-based centrality, PageRank variants, betweenness, closeness, and diffusion-based scores~\cite{CHEN2022613}.
However, most of these investigations focus on domains where the graph structure is readily available, such as social networks~\cite{analytics3040027}, citation graphs, biological interaction networks, and molecular graphs~\cite{ijcai2025p399}. In contrast, their application to image datasets remains limited, as graphs derived from visual data are typically constructed from feature similarity, where the interpretation of node importance differs. As a result, despite their potential, alternative centrality-based normalization strategies remain relatively underexplored in GNNs for image classification.

Motivated by this gap, this paper proposes the \textbf{G}aussian \textbf{Ra}nk-based \textbf{N}eighborhood \textbf{De}gree (\textbf{GRaNDe}), which incorporates neighborhood ranking and Gaussian weighting over distances to define node importance beyond simple connectivity.   Unlike the standard degree, by applying a Gaussian function to ranked distances, the proposed measure emphasizes closer and more relevant neighbors while smoothly attenuating the influence of more distant ones.  Also, GRaNDe can be easily integrated into existing GNN architectures as a direct replacement for degree-based normalization, improving information propagation and, consequently, effectiveness in image classification. The \textbf{main contributions} are summarized as follows:
\begin{itemize}    
    \item We propose GRaNDe, a novel weighted degree function to replace the standard degree centrality in GNNs.
    Unlike most existing approaches, which treat all neighbors of a node equally, GRaNDe incorporates distance information into the node degree through a Gaussian function and can be easily integrated into existing GNN models.

    \item Distances are updated with a Gaussian kernel during training, making GRaNDe adaptive by adjusting neighborhood importance as node representations evolve.

    \item Experiments on five public image classification datasets reveal that GRaNDe consistently improves classification accuracy across different settings, both on SGC and APPNP models.
    
    \item The proposed approach allows APPNP to sustain accuracy improvements as the number of neurons increases, effectively mitigating the performance saturation commonly observed in baseline approaches.
    
    \item A comparison with recent baselines shows that GRaNDe achieves superior or competitive results.
\end{itemize}

The remainder of this paper is organized as follows.
Section~\ref{sec:proposed_approach} details the main steps of our proposed approach, introduces GRaNDe, and explains how it is integrated into GNN models.
Section~\ref{sec:experiments} presents the experimental evaluation.
Finally, Section~\ref{sec:conclusions} concludes the paper.

\section{Proposed Approach}
\label{sec:proposed_approach}

\begin{figure*}[!t]
    \centering
    \makebox[\textwidth]{\includegraphics[width=1.0\textwidth]{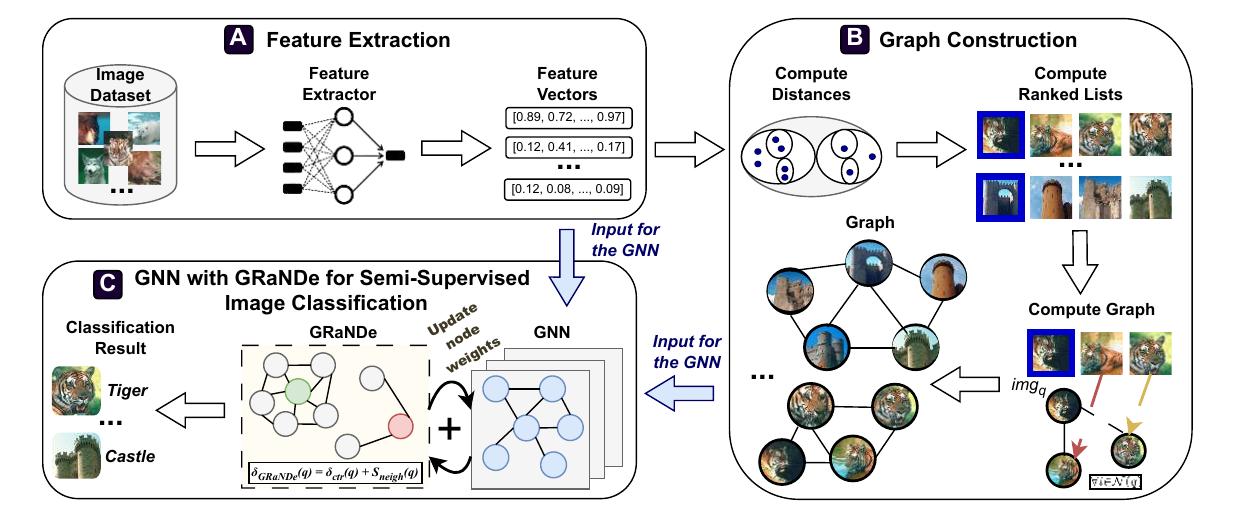}}
    \caption{Main steps of our GNN workflow with GRaNDe for semi-supervised image classification.}
    \label{fig:workflow}
\end{figure*}

Our proposed approach is composed of three main steps, illustrated in Figure~\ref{fig:workflow}, with GNN training incorporating GRaNDe as the central contribution. The following subsections detail each step depicted in the figure (A, B, and C).

\subsection{Feature Extraction}

Since images are inherently high-dimensional, feature extraction is employed to obtain compact and discriminative representations that facilitate the separation of samples in the feature space.
Traditionally, features were handcrafted by modeling visual characteristics such as color, shape, and texture. However, these descriptors are often limited in their ability to capture complex and high-level semantic patterns.

Current approaches rely on deep learning feature extractors, which generate more discriminative representations through transfer learning from models pretrained on large-scale datasets. In this work, we consider both Convolutional Neural Networks (CNNs) and Vision Transformers (ViTs), extracting image embeddings from the final fully connected layer in CNNs and from the CLS token representation in ViTs.

An image dataset can be defined as $\mathcal{C} = \{x_1, x_2, \dots, x_n\}$, where each image~$x_i$ is represented by a feature vector $\mathbf{x}_i \in \mathbb{R}^d$. The collection of all feature vectors is denoted by $\mathcal{X} = \{\mathbf{x}_1, \mathbf{x}_2, \dots, \mathbf{x}_n\} \subset \mathbb{R}^d$.
These feature vectors can be arranged into a matrix $\mathbf{X} \in \mathbb{R}^{n \times d}$, where each row corresponds to a feature vector $\mathbf{x}_i$ associated with an image~$x_i$.

\subsection{Graph Construction}

Beyond feature representations, the input graph is also a key component of the GNN learning process.
Firstly, we employ a BallTree algorithm to compute the Euclidean distances between feature vectors. By ordering these distances in ascending order, we obtain a ranked list of neighbors for each image. The ranked lists reflect the proximity between samples in the feature space, where closer neighbors are more similar and therefore more reliable for graph construction.

In this study, we consider undirected graphs $G = (V, E)$ built from a dataset of $n$ images, where each node represents a single image and the node set is defined as $V = \mathcal{X} = \{\mathbf{x}_1, \dots, \mathbf{x}_n\}$. In particular, we adopt the Reciprocal $k$-Nearest Neighbor Graph (Rec. $k$NN) commonly used in the literature~\cite{reID25, manifoldGCN} that relies on neighborhood information derived from ranked lists.

An edge between two images is created only if they appear in each other's $k$ nearest neighbor sets, a constraint that naturally provides a sparse graph, keeping only mutual edges:

\begin{equation}
E = \{(i, j) \;|\; j \in \mathcal{N}(\mathbf{x}_i, k) \;\wedge\; i \in \mathcal{N}(\mathbf{x}_j, k)\}.
\end{equation}

\subsection{GNN with GRaNDe for Image Classification}

Once the data are organized as a graph, learning can be formulated as the problem of propagating and refining information across related samples.
GNNs are designed to learn new node representations by jointly modeling node features and the underlying graph structure through neighborhood aggregation. These representations can be directly employed for classification at the node, edge, or graph level. In this work, we focus on node classification, where each node corresponds to an image.

Given the feature matrix $\mathbf{X}$ obtained in Step A and the graph $G$ constructed in Step B, the learning stage consists of training a GNN to propagate information over the graph and produce predictions.
This work considers diffusion-based GNNs, since they offer a simple and efficient formulation for feature propagation while preserving the structural information encoded in the graph. In these models, propagation is typically implemented through iterative multiplication by a normalized adjacency matrix $\mathbf{\hat{A}}$, where normalization is computed from the node degree.
In this step, we employ the proposed GRaNDe to guide the training process and perform node classification.

Subsection (1) describes the computation of the normalized adjacency matrix $\mathbf{\hat{A}}$. Subsections (2) and (3) present the two diffusion-based GNN models considered in this work and detail how $\mathbf{\hat{A}}$ is incorporated into their formulations. Finally, Subsection (4) introduces the proposed GRaNDe.

\subsubsection{Computing the Normalized Adjacency Matrix $\mathbf{\hat{A}}$}

Given an undirected graph $G = (V, E)$ with $n = |V|$ nodes, its structure can be represented by an adjacency matrix $\mathbf{A} \in \{0,1\}^{n \times n}$, where $\mathbf{A}_{ij} = 1$ if there exists an edge between nodes $v_i$ and $v_j$, and $\mathbf{A}_{ij} = 0$ otherwise. To allow each node to retain its own information during the propagation process, self-loops are added to the graph, producing the augmented adjacency matrix $\mathbf{\tilde{A}} = \mathbf{A} + \mathbf{I}$, where $\mathbf{I} \in \mathbb{R}^{n \times n}$ denotes the identity matrix.

The corresponding degree matrix $\mathbf{\tilde{D}} \in \mathbb{R}^{n \times n}$ is defined as a diagonal matrix whose entries are given by $\mathbf{\tilde{D}}_{ii} = \delta(i)$, where $\delta(i)$ denotes the node degree function, originally defined as the degree centrality $\delta_{\text{ctr}}$. This degree is computed as:
\begin{equation}
\delta_{\text{ctr}}(i) = \lvert \mathcal{N}(i) \rvert,
\end{equation}

\noindent where $\mathcal{N}(i)$ denotes the neighborhood of node $v_i$, including self-loops.
Therefore, $\lvert \mathcal{N}(i) \rvert$ is the number of neighbors.

The normalized adjacency matrix used for diffusion is then computed as $\mathbf{\hat{A}} = \mathbf{\tilde{D}}^{-\frac{1}{2}} \mathbf{\tilde{A}} \mathbf{\tilde{D}}^{-\frac{1}{2}}$.
This normalization rescales message propagation according to node connectivity, attenuating the influence of highly connected nodes while amplifying that of sparsely connected ones. As a result, $\mathbf{\hat{A}}$ defines a diffusion operator that enables stable and efficient information propagation across the graph, while preserving its structural properties. Repeated multiplication by $\mathbf{\hat{A}}$ corresponds to performing multi-step diffusion over the graph, which constitutes the core mechanism of diffusion-based GNN models adopted in this work.

\subsubsection{Simplified Graph Convolution (SGC)}

SGC~\cite{paperGCN-SGC2019} is a widely adopted GCN variant known for its lightweight architecture and computational efficiency, while keeping effective results in many cases.

Given the input feature matrix $\mathbf{X} \in \mathbb{R}^{n \times d}$, SGC first applies a linear transformation to the node features:
\begin{equation}
\mathbf{H}^{(0)} = \mathbf{X}\mathbf{W},
\end{equation}
where $\mathbf{W} \in \mathbb{R}^{d \times c}$ is a learnable weight matrix and $c$ denotes the number of classes.

The transformed features are then propagated through the graph by performing $K$ successive diffusion steps using the normalized adjacency matrix $\mathbf{\hat{A}}$:
\begin{equation}
\mathbf{Z} = \mathbf{\hat{A}}^{K} \mathbf{H}^{(0)}.
\end{equation}

\noindent The resulting representations $\mathbf{Z}$ are used for classification.

\subsubsection{Approximate Personalized Propagation of Neural Predictions (APPNP)}

APPNP~\cite{paperGCN-APPNP2019} is a diffusion-based GNN model inspired by the Personalized PageRank formulation. Unlike SGC, APPNP decouples feature transformation from propagation and employs a teleport mechanism to preserve the original node representations during diffusion, alleviating the over-smoothing effect commonly observed in deep propagation.

Given the input feature matrix $\mathbf{X} \in \mathbb{R}^{n \times d}$, APPNP first applies a linear transformation:
$\mathbf{H}^{(0)} = \mathbf{X}\mathbf{W}$,
where $\mathbf{W} \in \mathbb{R}^{d \times c}$ is a learnable weight matrix and $c$ denotes the number of classes.
The transformed features are then propagated through the graph using an iterative diffusion process defined as:
\begin{equation}
\mathbf{H}^{(k)} = (1 - \alpha)\mathbf{\hat{A}} \mathbf{H}^{(k-1)} + \alpha \mathbf{H}^{(0)}, \quad k = 1, \dots, K,
\end{equation}
where $\alpha \in (0,1)$ is the teleport probability that controls the trade-off between propagating information from neighboring nodes and retaining the original node representations.

After $K$ propagation steps, the final node representations are given by $\mathbf{Z} = \mathbf{H}^{(K)}$.
By repeatedly diffusing predictions while injecting the initial representations at each step, APPNP implements an approximation of Personalized PageRank over the graph. This mechanism enables effective long-range information propagation while maintaining stability and preserving discriminative node features.

\subsubsection{Gaussian Rank-Based Neighborhood Degree (GRaNDe)}

The proposed degree function is used as a direct replacement for the standard degree centrality in both SGC and APPNP models, so that the node degree $\delta(i)$ is no longer defined as $\delta_{\text{ctr}}$, but as the proposed $\delta_{\text{GRaNDe}}$.
GRaNDe assigns a score to each node $q$, which is used to modulate message amplification during GNN propagation, with lower values leading to a stronger influence. GRaNDe is defined as:
\begin{equation}
    \delta_{\text{GRaNDe}}(q) = \delta_{\text{ctr}}(q) + s_{\text{neigh}}(q),
\end{equation}
\noindent where $\delta_{\text{ctr}}(q)$ denotes the degree centrality of node $q$, and $s_{\text{neigh}}(q)$ is a neighborhood penalty term.

By using $\delta_{\text{ctr}}(q)$ as part of $\delta_{\text{GRaNDe}}(q)$, we preserve the original intuition behind degree normalization, that nodes with many connections should have their messages attenuated to avoid dominating the propagation process and causing excessive smoothing of representations. In contrast, nodes with few connections are assigned smaller degree values, which amplifies their messages and allows them to exert a stronger influence during information diffusion.

On the other hand, $s_{\text{neigh}}(q)$ acts as a penalty term that measures the average dissimilarity between an image $q$ and its neighbors.
Specifically, it computes the mean of the inverse Gaussian similarities derived from the distances between $q$ and the samples in its neighborhood, where the averaging factor corresponds to the neighborhood size (i.e., $\delta_{\text{ctr}}(q) = \lvert \mathcal{N}(q) \rvert$). Consequently, larger values of $s_{\text{neigh}}(q)$ indicate neighborhoods whose samples are farther from $q$ on average, while smaller values correspond to closer neighborhoods.
Therefore, we define the neighborhood penalty score as:

\begin{equation}
\label{eq:penalty}
s_{\text{neigh}}(q)=\frac{1}{\delta_{\text{ctr}}(q)}\sum_{i\in\mathcal{N}(q)}
\left[f_g(q,i,\sigma)\right]^{-1},
\end{equation}

\noindent where the function $f_{g}(q,i,\sigma)$ corresponds to a Gaussian radial basis function (RBF) applied to the Euclidean distance between nodes $q$ and $i$.

The distances used by $f_g$ are computed in the feature space that is progressively updated during training. As a consequence, the neighborhood penalty term $s_{\text{neigh}}(q)$ is recomputed at each training epoch using the updated feature representations $\mathbf{H}^{(0)}$. This allows GRaNDe to operate as a dynamic degree function, adapting to the evolving geometry of the feature space learned.
Let $\mathbf{h}_q$ and $\mathbf{h}_i$ denote the feature representations of nodes $q$ and $i$, respectively. The pairwise distance between nodes is computed using the Euclidean ($\ell_2$) norm,
$\rho(q,i) = \lVert \mathbf{h}_q - \mathbf{h}_i \rVert_2 $.
All distances are min-max normalized, since the Gaussian kernel is sensitive to the input range, and the resulting value is denoted by $\rho'(q,i) \in [0,1]$.
The Gaussian kernel function is defined as:
\begin{equation}
f_{g}(q,i,\sigma) = \exp\left( -\frac{\rho'(q,i)^2}{\sigma} \right),
\end{equation}
\noindent where the parameter $\sigma > 0$ is a positive scaling parameter that controls the rate at which the influence of a node $i$ decays as its distance to $q$ increases.
The more similar the nodes, the higher the value.
For notational simplicity, we also write the kernel as $f_g(\rho')$.

Please note that the inverse term in Equation~\ref{eq:penalty} is intentionally used so that nodes with distant neighbors receive higher penalties, and vice versa.
Moreover, division by zero is not an issue, as the Gaussian RBF is strictly positive for any finite distance, and the degree $\delta_{\text{ctr}}$ is always greater than or equal to one for any connected node, due to the presence of a self-loop at each node.

\begin{figure}[!ht]
    \centering
    \includegraphics[width=\textwidth]{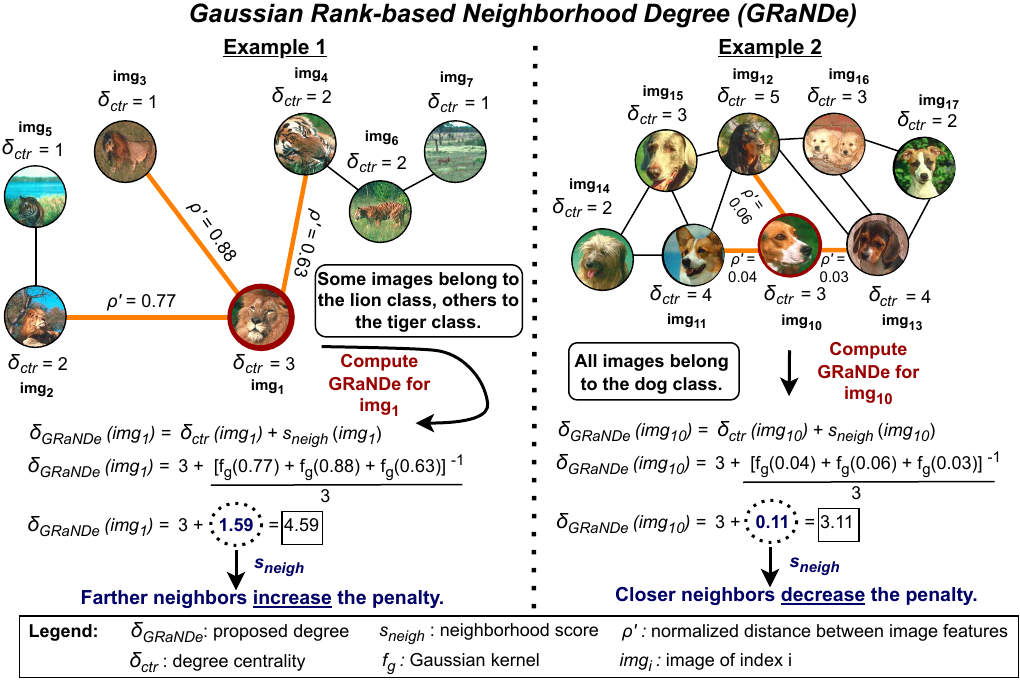}
    \caption{Illustration that exemplifies how GRaNDe is computed in two scenarios: \emph{(i)} a node with farther, sparsely connected neighbors and \emph{(ii)} a node with closer, densely connected neighbors. Images are from the Corel5k dataset. Self-loops are omitted for visualization purposes.}
    \label{fig:grande}
\end{figure}

Figure~\ref{fig:grande} illustrates examples of how GRaNDe is computed. In example 1, the central node (in red) is connected to farther neighbors, leading to smaller Gaussian similarity values, which results in a higher neighborhood penalty and a larger GRaNDe value. In example 2, the node is surrounded by closer neighbors, providing higher similarity scores, reducing the penalty term, and producing a smaller GRaNDe value.

\section{Experimental Evaluation}
\label{sec:experiments}

\subsection{Experimental Setup}

The evaluation was conducted on five publicly available image datasets, as follows:

\begin{itemize}
  \item \textbf{Flowers17}~\cite{PaperFlowers}: contains 1,360 images evenly distributed across 17 flower categories. Each class includes 80 images captured under diverse backgrounds and viewpoints, resulting in moderate intra-class variability.
  
  \item \textbf{Corel5k}~\cite{PaperCorel5k_PR2013}: consists of 5,000 natural images spanning 50 semantic categories, including animals, landscapes, and indoor and outdoor scenes.
  
  \item \textbf{Pets}~\cite{parkhi12a}: comprises 7,409 images from 37 pet classes, with significant variability in pose, illumination, scale, and background.

  \item \textbf{CUB200}~\cite{WahCUB_200_2011}: includes 11,788 images of 200 bird species, providing detailed annotations commonly used for fine-grained recognition and localization tasks.

  \item \textbf{Dogs}~\cite{paperDogs}: contains 20,580 images of 120 dog breeds, presenting substantial intra-class variation and inter-class similarity, which makes it a challenging benchmark for fine-grained image classification.
\end{itemize}

For every dataset, experiments were conducted using four deep learning models pretrained on the ImageNet dataset:

\begin{itemize}
  \item \textbf{ResNet152}~\cite{paperRESNET}: a deep residual CNN. We used the 2048-dimensional output from the final convolutional stage.

  \item \textbf{SENet154}~\cite{paperCNN_SENET_2018}: a ResNet variant with channel-wise attention. Features were extracted from the final SE block (2048 dimensions).

   \item \textbf{DPNet92}~\cite{paperCNN_DPN2017}: a dual-path CNN that combines residual feature reuse with dense feature exploration. We used the global pooled output from the final stage (2688 dimensions).

    \item \textbf{ViT-B16}~\cite{paperVIT16}: a transformer model operating on patch sequences. We used the 768-dimensional class token from the last encoder layer.
\end{itemize}

Regarding the setup, we considered both SGC and APPNP models.
In all experiments, we considered a reciprocal kNN graph with $k = 40$. We employed a learning rate of $10^{-3}$ for all datasets, except for CUB200, where a learning rate of $10^{-2}$ was used.
For the GNN hyperparameters, we followed the default settings provided by the PyTorch Geometric repository\footnote{\url{www.github.com/pyg-team/pytorch_geometric}}, using 256 neurons as the hidden dimension for APPNP.
All GNNs were trained for 200 epochs in each execution using the Adam optimizer.
All results are reported using a 10-fold protocol, in which each fold is used once as the training set while the remaining folds are used for testing, following a setup with 10\% of the data for training and 90\% for testing with the objective of evaluating the proposed method in a semi-supervised setting with limited labeled data. Each reported result corresponds to the mean accuracy over 5 executions of this 10-fold procedure, and the standard deviation is reported to show accuracy variability.
These settings were chosen to ensure a fair comparison, consistent with the settings used by other methods.

\subsection{Analysis of the Sigma ($\sigma$) Parameter}

\begin{figure}[th!]
    \centering
    \includegraphics[width=0.56\textwidth]{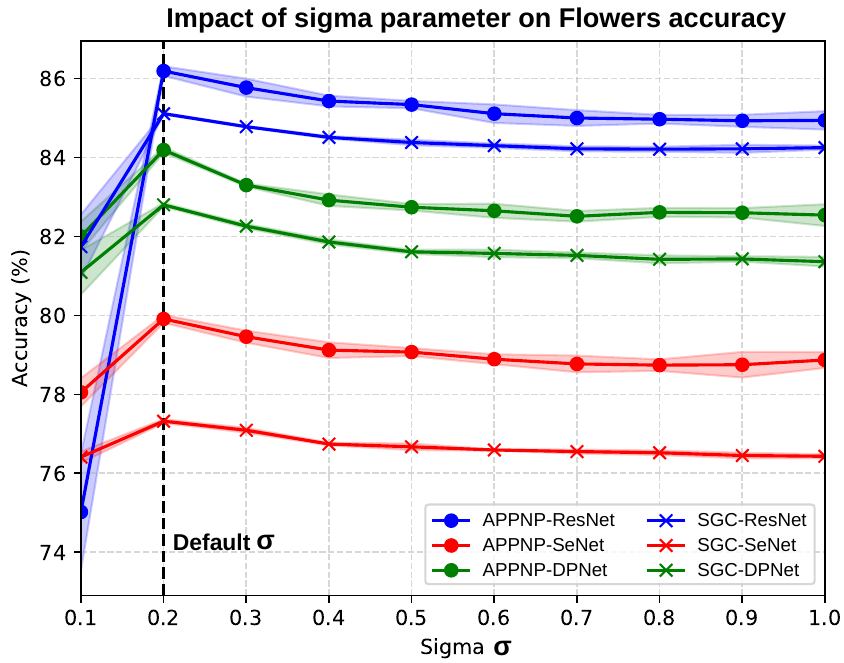}
    \caption{Impact of the $\sigma$ parameter on classification accuracy for Flowers.}
    \label{fig:sigma}
\end{figure}

The parameter $\sigma$ plays a crucial role in the Gaussian function $f_g$, as it controls the sensitivity of the kernel to distance variations, determining how quickly similarity decays as samples move farther apart in the feature space.
Figure~\ref{fig:sigma} illustrates the effect of the $\sigma$ parameter on classification accuracy for different feature extractors using both SGC and APPNP models on the Flowers dataset. The shaded area indicates the standard deviation over multiple runs. Overall, $\sigma = 0.2$ achieved the highest accuracy in most configurations. Therefore, we set $\sigma = 0.2$ as the default parameter.

\subsection{Semi-Supervised Image Classification}

\begin{table*}[!t]
\caption{Accuracy (\%) gains for APPNP and SGC models when GRaNDe is employed. Results are reported for the default and best sigma.
For each row, the best result is highlighted in bold. The best result for each dataset and GNN is highlighted in blue.}
\label{tab:grande_gains}
\centering
\resizebox{.84\textwidth}{!}{
\begin{tabular}{l|l|l|c|c|cc|l}
\hline
\textbf{GNN} & \textbf{Dataset} & \textbf{Features} & \textbf{Accuracy with} & \multicolumn{3}{c|}{\cellcolor{cinza}\textbf{Accuracy with GRaNDe (ours)}} & \textbf{Relative} \\
\cline{5-7}
 & & & \textbf{Degree Centrality} & \textbf{Default $\sigma=0.2$} & \textbf{Best $\sigma$} & \textbf{$\sigma$} &  ~~\textbf{Gain} \\
\hline
\hline

\multirow{25}{*}{\rotatebox{45}{\textbf{APPNP}}}
& \multirow{4}{*}{\textbf{Flowers17}}
 & \textbf{ResNet}  & 84.43 $\pm$ 0.088 & {\textbf{86.19 $\pm$ 0.121}} & $86.19 \pm 0.121$ & 0.2 &  {\color{DarkGreen}\textbf{+2.09\%}} \\
&  & \textbf{SENet}   & 77.88 $\pm$ 0.190 & {\textbf{79.91 $\pm$ 0.097}} & 79.91 $\pm$ 0.097 & 0.2 &  {\color{DarkGreen}\textbf{+2.61\%}} \\
&  & \textbf{DPNet}   & 82.00 $\pm$ 0.153 & {\textbf{84.19 $\pm$ 0.070}} & 84.19 $\pm$ 0.070 & 0.2 &  {\color{DarkGreen}\textbf{+2.67\%}} \\
&  & \textbf{ViT-B16} & 97.14 $\pm$ 0.019 & {\color{DarkBlue}\textbf{98.18 $\pm$ 0.048}} & 98.18 $\pm$ 0.048 & 0.2 &  {\color{DarkGreen}\textbf{+1.08\%}} \\
\cline{2-8}

& \multirow{4}{*}{\textbf{Corel5K}}
 & \textbf{ResNet}  & 92.65 $\pm$ 0.073 & {\textbf{93.25 $\pm$ 0.040}} & 93.25 $\pm$ 0.040 & 0.2 &  {\color{DarkGreen}\textbf{+0.65\%}} \\
&  & \textbf{SENet}   & 92.63 $\pm$ 0.066 & {\textbf{93.42 $\pm$ 0.050}} & 93.42 $\pm$ 0.050 & 0.2 &  {\color{DarkGreen}\textbf{+0.85\%}} \\
&  & \textbf{DPNet}   & 90.35 $\pm$ 0.079 & \textbf{91.75 $\pm$ 0.040} & 91.75 $\pm$ 0.040 & 0.2 &  {\color{DarkGreen}\textbf{+1.56\%}} \\
&  & \textbf{ViT-B16} & 94.24 $\pm$ 0.171 & 89.01 $\pm$ 0.304 & {\color{DarkBlue}\textbf{94.64 $\pm$ 0.096}} & 0.3 &  {\color{DarkGreen}\textbf{+0.42\%}} \\
\cline{2-8}

& \multirow{4}{*}{\textbf{CUB200}}
 & \textbf{ResNet}  & 49.49 $\pm$ 0.960 & {\textbf{54.58 $\pm$ 0.238}} & 54.58 $\pm$ 0.238 & 0.2 &  {\color{DarkGreen}\textbf{+10.27\%}} \\
&  & \textbf{SENet}   & 38.08 $\pm$ 0.065 & {\textbf{43.78 $\pm$ 0.178}} & 43.78 $\pm$ 0.178 & 0.2 &  {\color{DarkGreen}\textbf{+12.03\%}} \\
&  & \textbf{DPNet}   & 39.94 $\pm$ 0.507 & 41.69 $\pm$ 0.690 & \textbf{46.32 $\pm$ 0.257} & 0.3 &  {\color{DarkGreen}\textbf{+15.98\%}} \\
&  & \textbf{ViT-B16} & 71.56 $\pm$ 0.230 & 65.23 $\pm$ 0.607 & {\color{DarkBlue}\textbf{74.41 $\pm$ 0.222}} & 0.4 & {\color{DarkGreen}\textbf{+3.99\%}} \\
\cline{2-8}

& \multirow{4}{*}{\textbf{Pets}}
 & \textbf{ResNet}    & 90.06 $\pm$ 0.051 & \textbf{91.20 $\pm$ 0.042} &  91.20 $\pm$ 0.042  & 0.2 &  {\color{DarkGreen}\textbf{+1.26\%}}\\
&  & \textbf{SENet}   & 89.69 $\pm$ 0.051 & \textbf{90.56 $\pm$ 0.037} &  90.56 $\pm$ 0.037  & 0.2 &  {\color{DarkGreen}\textbf{+0.97\%}}\\
&  & \textbf{DPNet}   & 91.13 $\pm$ 0.068 & {\color{DarkBlue}\textbf{91.78 $\pm$ 0.041}} &  91.78 $\pm$ 0.041  & 0.2 &  {\color{DarkGreen}\textbf{+0.72\%}}\\
&  & \textbf{ViT-B16} & 87.83 $\pm$ 0.036 & 86.37 $\pm$ 0.168 & \textbf{88.50 $\pm$ 0.039} & 0.3 &    {\color{DarkGreen}\textbf{+0.76\%}}\\
\cline{2-8}

& \multirow{4}{*}{\textbf{Dogs}}
 & \textbf{ResNet}    & 88.46 $\pm$ 0.039 & 89.12 $\pm$ 0.023 & \textbf{89.25 $\pm$ 0.023} & 0.3 &  {\color{DarkGreen}\textbf{+0.90\%}}\\
&  & \textbf{SENet}   & 94.13 $\pm$ 0.014 & {\color{DarkBlue}\textbf{94.31 $\pm$ 0.013}} & 94.31 $\pm$ 0.013 & 0.2 &  {\color{DarkGreen}\textbf{+0.19\%}}\\
&  & \textbf{DPNet}   & 92.65 $\pm$ 0.050 & 91.68 $\pm$ 0.180 & \textbf{93.05 $\pm$ 0.019} & 0.1 &  {\color{DarkGreen}\textbf{+0.43\%}}\\
&  & \textbf{ViT-B16} & 91.00 $\pm$ 0.012 & 76.83 $\pm$ 0.136 & \textbf{91.34 $\pm$ 0.011} & 0.4 &  {\color{DarkGreen}\textbf{+0.38\%}}\\
\hline
\hline

\multirow{25}{*}{\rotatebox{45}{\textbf{SGC}}}
& \multirow{4}{*}{\textbf{Flowers17}}
 & \textbf{ResNet}  & 83.94 $\pm$ 0.041 & \textbf{85.11 $\pm$ 0.021} & 85.11 $\pm$ 0.021 & 0.2 &  {\color{DarkGreen}\textbf{+1.39\%}} \\
&  & \textbf{SENet}   & 76.17 $\pm$ 0.051 & \textbf{77.32 $\pm$ 0.048} & 77.32 $\pm$ 0.048 & 0.2 &  {\color{DarkGreen}\textbf{+1.51\%}} \\
&  & \textbf{DPNet}   & 81.30 $\pm$ 0.095 & \textbf{82.80 $\pm$ 0.046} & 82.80 $\pm$ 0.046 & 0.2 &  {\color{DarkGreen}\textbf{+1.84\%}} \\
&  & \textbf{ViT-B16} & 96.92 $\pm$ 0.034 & 97.38 $\pm$ 0.024 & {\color{DarkBlue}\textbf{97.76 $\pm$ 0.014}} & 0.1 &  {\color{DarkGreen}\textbf{+0.86\%}} \\
\cline{2-8}

& \multirow{4}{*}{\textbf{Corel5K}}
 & \textbf{ResNet}  & 91.98 $\pm$ 0.006 & \textbf{92.55 $\pm$ 0.008} & 92.55 $\pm$ 0.008 & 0.2 &  {\color{DarkGreen}\textbf{+0.62\%}} \\
&  & \textbf{SENet}   & 92.15 $\pm$ 0.014 & 92.18 $\pm$ 0.035 & \textbf{92.76 $\pm$ 0.012} & 0.1 &  {\color{DarkGreen}\textbf{+0.66\%}} \\
&  & \textbf{DPNet}   & 89.78 $\pm$ 0.042 & {\textbf{91.83 $\pm$ 0.043}} & 91.83 $\pm$ 0.043 & 0.2 &  {\color{DarkGreen}\textbf{+2.28\%}} \\
&  & \textbf{ViT-B16} & 95.47 $\pm$ 0.057 & {\color{DarkBlue}\textbf{95.77 $\pm$ 0.015}} & 95.77 $\pm$ 0.015 & 0.2 &  {\color{DarkGreen}\textbf{+0.31\%}} \\
\cline{2-8}

& \multirow{4}{*}{\textbf{CUB200}}
 & \textbf{ResNet}  & 53.79 $\pm$ 0.019 & 43.58 $\pm$ 0.614 & \textbf{54.45 $\pm$ 0.027} & 0.4 &  {\color{DarkGreen}\textbf{+1.23\%}} \\
&  & \textbf{SENet}   & 40.32 $\pm$ 0.011 & 41.29 $\pm$ 0.007 & \textbf{41.79 $\pm$ 0.013} & 0.3 &  {\color{DarkGreen}\textbf{+3.64\%}} \\
&  & \textbf{DPNet}   & 52.48 $\pm$ 0.578 & 52.46 $\pm$ 0.042 & {\textbf{52.69 $\pm$ 0.052}} & 0.7 &  {\color{DarkGreen}\textbf{+0.40\%}} \\
&  & \textbf{ViT-B16} & 78.35 $\pm$ 0.030 & 77.92 $\pm$ 0.020 & {\color{DarkBlue}\textbf{78.58 $\pm$ 0.013}} & 0.4 &  {\color{DarkGreen}\textbf{+0.30\%}} \\
\cline{2-8}

& \multirow{4}{*}{\textbf{Pets}}
 & \textbf{ResNet}  & 89.85 $\pm$ 0.016 & 90.45 $\pm$ 0.019 & \textbf{90.79 $\pm$ 0.004} & 0.1 &  {\color{DarkGreen}\textbf{+1.04\%}}\\
&  & \textbf{SENet}   & 89.65 $\pm$ 0.017& 89.87 $\pm$ 0.014& \textbf{89.98 $\pm$ 0.017} & 0.1 &  {\color{DarkGreen}\textbf{+0.37\%}} \\
&  & \textbf{DPNet}   & 91.08 $\pm$ 0.034 & {\color{DarkBlue}\textbf{91.29 $\pm$ 0.024}} & 91.29 $\pm$ 0.024 & 0.2 &  {\color{DarkGreen}\textbf{+0.23\%}}\\
&  & \textbf{ViT-B16} & 88.22 $\pm$ 0.032 & \textbf{88.34 $\pm$ 0.023} & 88.34 $\pm$ 0.023 & 0.2 &  {\color{DarkGreen}\textbf{+0.14\%}} \\
\cline{2-8}

& \multirow{4}{*}{\textbf{Dogs}}
 & \textbf{ResNet}    & 88.55 $\pm$ 0.014 & \textbf{89.03 $\pm$ 0.010} & 89.03 $\pm$ 0.010 & 0.2  &  {\color{DarkGreen}\textbf{+0.54\%}} \\
&  & \textbf{SENet}   & 94.32 $\pm$ 0.008 & 94.25 $\pm$ 0.007 & \color{DarkBlue}{\textbf{94.35 $\pm$ 0.007}} & 0.1 &  {\color{DarkGreen}\textbf{+0.04\%}} \\
&  & \textbf{DPNet}   & 93.10 $\pm$ 0.009 & {93.00 $\pm$ 0.010} & \textbf{93.13 $\pm$ 0.009} & 0.3 &  {\color{DarkGreen}\textbf{+0.03\%}} \\
&  & \textbf{ViT-B16} & 91.76 $\pm$ 0.010 & {\textbf{91.91 $\pm$ 0.009}} & 91.91 $\pm$ 0.009 & 0.2 &  {\color{DarkGreen}\textbf{+0.16\%}}\\
\hline

\end{tabular}
}
\end{table*}

We conducted experiments on five datasets using four different feature extractors, replacing the standard degree centrality $\delta_{\text{ctr}}$ with the proposed GRaNDe ($\delta_{\text{GRaNDe}}$), and compared the accuracy.
Table~\ref{tab:grande_gains} reports the results obtained with standard degree centrality and with GRaNDe. 
For our method, we report results using the default value $\sigma = 0.2$, as well as the $\sigma$ that achieved the highest accuracy for each setting. The parameter $\sigma$ was evaluated over the range $[0.1, 1.0]$ with a step size of $0.1$.
Relative gains are computed by comparing the results with and without GRaNDe.

Overall, GRaNDe consistently improves
accuracy in both GNNs across all evaluated datasets and features.
Notice that the highest gains are observed for APPNP, particularly on the CUB200 dataset, with DPNet (+15.98\%) and SENet (+12.03\%).
 This is probably because APPNP performs multiple propagation steps, which progressively reinforce the influence of GRaNDe throughout the diffusion process. In contrast, SGC relies on a single linear diffusion step, limiting the accumulation of such effects.

\subsection{Comparison with Other Approaches}

Since APPNP achieved higher relative gains in most evaluated configurations, we further extended its evaluation by comparing it against recent baselines using different numbers of neurons.
We considered both Manifold-GCN (MGCN)~\cite{manifoldGCN} and Density-Guided Correlation Graph (DGCG)~\cite{DGCG} under the same experimental settings and hyperparameters by using the publicly available implementations.

Both baselines use GNNs for semi-supervised image classification.
Manifold-GCN leverages manifold learning to build graphs that capture the intrinsic geometric structure of the feature space for semi-supervised classification, while DGCG builds graphs using rank correlation measures and an adaptive thresholding strategy guided by graph density.

Table~\ref{tab:appnp_neurons} reports a comparison of our approach using APPNP against competing methods under different hidden layer sizes (64, 128, and 256 neurons). For a fixed dataset and feature extractor, GRaNDe consistently achieves the best accuracy across most configurations.
Moreover, while the baseline methods do not consistently benefit from larger hidden layers in all cases, GRaNDe shows accuracy improvements in most cases when the number of neurons is increased.

\begin{table}[!ht]

\centering
\caption{\centering {Impact of the number of neurons on APPNP for our approach.}}
\label{tab:appnp_neurons}
\resizebox{.54\textwidth}{!}{
\begin{tabular}{l|l|c|cc|cc}
\hline
\multirow{2}{*}{\textbf{Dataset}} & \multirow{2}{*}{\textbf{Feature}} & \textbf{Num. of} & \multirow{2}{*}{\textbf{MGCN}} & \multirow{2}{*}{\textbf{DGCG}} & \cellcolor{cinza}\multirow{1}{*}{\textbf{GRaNDe}} & \cellcolor{cinza}\multirow{1}{*}{\textbf{$\sigma$}}
 \\
 & & \textbf{Neurons} &  &  & \textbf{(ours)}\cellcolor{cinza} & \cellcolor{cinza} 
 \\
\hline \hline
\multirow{9}{*}{\rotatebox{45}{\textbf{Flowers}}} 
 & \multirow{3}{*}{\textbf{ResNet}}  & 64  & 85.41 & 85.08 & 84.11 & \multirow{3}{*}{0.2} \\
 &                                   & 128 & 85.38 & 85.06 & 85.81 \\
 &                                   & 256 & 85.35 & 85.14 & \textbf{86.19} \\
\cline{2-7}
 & \multirow{3}{*}{\textbf{SENet}}   & 64  & 78.82 & 77.40 & 79.90 &\multirow{3}{*}{0.2} \\
 &                                   & 128 & 78.72 & 77.03 & \textbf{80.10} \\
 &                                   & 256 & 78.59 & 76.79 & 79.91 \\
\cline{2-7}
 & \multirow{3}{*}{\textbf{ViT-B16}} & 64  & 97.43 & 97.68 & 96.62 &\multirow{3}{*}{0.2} \\
 &                                   & 128 & 97.34 & 97.79 & 97.32 \\
 &                                   & 256 & 97.31 & 97.86 & \textbf{98.18} \\
\hline \hline
\multirow{9}{*}{\rotatebox{45}{\textbf{Corel5k}}} 
 & \multirow{3}{*}{\textbf{ResNet}}  & 64  & 92.82 & 90.68 & 83.33 &\multirow{3}{*}{0.2} \\
 &                                   & 128 & 92.72 & 91.10 & 91.74 \\
 &                                   & 256 & 92.54 & 91.27 & \textbf{93.25} \\
\cline{2-7}
 & \multirow{3}{*}{\textbf{SENet}}   & 64  & 91.92 & 90.46 & 80.80 &\multirow{3}{*}{0.2} \\
 &                                   & 128 & 91.79 & 90.75 & 90.61 \\
 &                                   & 256 & 91.73 & 90.86 & \textbf{93.42} \\
\cline{2-7}
 & \multirow{3}{*}{\textbf{ViT-B16}} & 64  & 95.13 & 90.23 & 92.79 &\multirow{3}{*}{0.3}  \\
 &                                   & 128 & 95.21 & 92.28 & 93.93 \\
 &                                   & 256 & \textbf{95.22} & 92.84 & 94.64 \\
\hline \hline
\multirow{9}{*}{\rotatebox{45}{\textbf{CUB200}}} 
 & \multirow{3}{*}{\textbf{ResNet}}  & 64  & 48.44 & 51.54 & 51.77 &\multirow{3}{*}{0.2}\\
 &                                   & 128 & 51.29 & 53.66 & 54.53 \\
 &                                   & 256 & 51.92 & 53.99 & \textbf{54.58} \\
\cline{2-7}
 & \multirow{3}{*}{\textbf{SENet}}   & 64  & 39.15 & 38.20 & 41.93 &\multirow{3}{*}{0.2} \\
 &                                   & 128 & 38.84  & 40.16 & 43.28 \\
 &                                   & 256 & 38.64 & 40.79 & \textbf{43.78} \\
\cline{2-7}
 & \multirow{3}{*}{\textbf{ViT-B16}} & 64  & 73.35 & 61.31 & 72.55 &\multirow{3}{*}{0.4} \\
 &                                   & 128 & 76.12 & 69.00 & 74.24 \\
 &                                   & 256 & \textbf{76.98} & 72.87 & 74.41 \\
\hline
\end{tabular}
}
\end{table}

We also conducted a broader comparison against a diverse set of image classification approaches, as reported in Table~\ref{tab:baselines_classification}. In this setting, the comparison is not limited to graph neural network-based approaches, but also includes a wider range of traditional and recent classification methods evaluated under the same experimental protocol. All methods were run using identical train/test splits and feature extractors to ensure a fair comparison.	In this table, for our approach as well as DGCG and MGCN, we report the best accuracy achieved across both SGC and APPNP.	Note that CoMatch~\cite{Li_2021_ICCV} operates directly on input images rather than extracted features and therefore provides the same accuracy regardless of the feature extractor. We also include the graph-based semi-supervised method GNN-LDS~\cite{pmlr-v97-franceschi19a}, as well as WSEF~\cite{paperWSEF}, which leverages weakly supervised embeddings.	Additionally, we included two commonly adopted semi-supervised strategies as baselines: label spreading (LS), which propagates labels to unlabeled samples before classification, and pseudolabeling (PL), a self-training approach that iteratively assigns pseudo-labels based on model predictions. The OPF classifier~\cite{paperSSOPF2014} is also considered in both its standard and label-propagation variants.	The abbreviation ML-prec denotes multi-layer perceptrons. When a reference is not provided, the method corresponds to a traditional approach from the \textit{sklearn} library.	Overall, GRaNDe achieved the best results in the majority of cases.

\begin{table*}[!ht]
  \centering
  \scriptsize
  \setlength{\tabcolsep}{3pt}
    \caption{Accuracy (\%) for semi-supervised image classification using the same features, with only 10\% training data. For each method, we report the best accuracy. Our results are in green, and the best per row is in bold.}
  \label{tab:baselines_classification}
  \resizebox{\textwidth}{!}{
  \begin{tabular}{l l|*{14}{c}}
    \hline
    \textbf{Dataset} & \textbf{Features}
      & \shortstack{\textbf{CoMatch}}
      & \textbf{SVM}
      & \shortstack{\textbf{OPF}}
      & \shortstack{\textbf{ML-}\\\textbf{Perc.}}
      & \shortstack{\textbf{PL+}\\\textbf{SGD}}
      & \shortstack{\textbf{LS+}\\\textbf{kNN}}
      & \shortstack{\textbf{LS+}\\\textbf{SVM}}
      & \shortstack{\textbf{LS+}\\\textbf{OPF}}
      & \shortstack{\textbf{LS+}\\\textbf{ML-}\\\textbf{Perc.}}
      & \shortstack{\textbf{GNN-}\\\textbf{LDS}}
      & \shortstack{\textbf{WSEF}}
      & \shortstack{\textbf{MGCN}}
      & \shortstack{\textbf{DGCG}}
      & \cellcolor{cinza}\shortstack{\textbf{GRaNDe}\\\textbf{(ours)}} \\
    \hline
    \multirow{3}{*}{\rotatebox{45}{\textbf{Flowers}}}
      & \textbf{ResNet}
        & \multirow{3}{*}{\rotatebox{45}{82.55}} & 80.54 & 71.77 & 78.88 & 82.69
        & 73.49 & 73.53 & 72.66 & 73.03
        & 79.32 & 85.12
        & 85.88
        & 85.68
        & \cellcolor{cinza}\textcolor{DarkGreen}{\textbf{86.19}} \\
      & \textbf{SENet}
        & & 73.30 & 64.00 & 72.62 & 76.87
        & 58.05 & 59.84 & 59.25 & 59.39
        & 73.69 & 76.16
        & 78.82
        & 78.42
        & \cellcolor{cinza}\textcolor{DarkGreen}{\textbf{80.10}} \\
      & \textbf{ViT-B16}
        &  & 96.75 & 96.50 & 92.59 & 96.84
        & 95.74 & 94.49 & 94.22 & 95.13
        & 96.66 & 97.82
        & 97.43
        & 98.09
        & \cellcolor{cinza}\textcolor{DarkGreen}{\textbf{98.18}} \\
    \hline
    \multirow{3}{*}{\rotatebox{45}{\textbf{Corel5k}}}
      & \textbf{ResNet}
        & \multirow{3}{*}{\rotatebox{45}{85.70}} & 88.73 & 83.56 & 87.10 & 89.76
        & 83.98 & 83.26 & 82.32 & 82.53
        & 88.94 & 91.68
        & 93.08
        & 91.48
        & \cellcolor{cinza}\textcolor{DarkGreen}{\textbf{93.25}} \\
      & \textbf{SENet}
        &  & 85.89 & 81.33 & 86.90 & 89.85
        & 72.16 & 72.79 & 72.20 & 72.24
        & 89.95 & 89.74
        & 92.79
        & 90.88
        & \cellcolor{cinza}\textcolor{DarkGreen}{\textbf{93.42}} \\
      & \textbf{ViT-B16}
        &  & 91.92 & 90.02 & 74.41 & 89.07
        & 89.63 & 87.59 & 86.14 & 87.68
        & 88.56 & 94.00
        & 95.57
        & 95.29
        & \cellcolor{cinza}\textcolor{DarkGreen}{\textbf{95.77}} \\
    \hline
    \multirow{3}{*}{\rotatebox{45}{\textbf{CUB200}}}
      & \textbf{ResNet}
        & \multirow{3}{*}{\rotatebox{45}{38.29}} & 48.84 & 38.59 & 32.24 & 21.67
        & 36.99 & 38.70 & 39.28 & 39.68
        & 37.78 & 52.17
        & 52.85
        & 54.43
        & \cellcolor{cinza}\textcolor{DarkGreen}{\textbf{54.58}} \\
      & \textbf{SENet}
        &  & 35.32 & 30.94 & 32.15 & 20.96
        & 20.00 & 24.82 & 25.38 & 25.72
        & --- & 36.49
        & 40.31
        & 41.00
        & \cellcolor{cinza}\textcolor{DarkGreen}{\textbf{43.78}} \\
      & \textbf{ViT-B16}
        &  & 75.61 & 73.27 & 12.02 & 30.19
        & 66.15 & 66.81 & 66.68 & 62.81
        & 52.42 & 78.64
        & 79.27
        & \textbf{81.23}
        & \cellcolor{cinza}\textcolor{DarkGreen}{78.58} \\
    \hline
  \end{tabular}
  }
\end{table*}

\section{Conclusion}
\label{sec:conclusions}

In this work, we introduced GRaNDe, a novel degree function for weighting nodes during GNN training. Many traditional diffusion-based GNNs rely on degree-based normalization, attenuating messages from highly connected nodes and amplifying those from sparsely connected ones, regardless of neighborhood quality. In contrast, GRaNDe extends this formulation by explicitly incorporating a neighborhood penalty score. 
Nodes whose neighbors are farther apart in the feature space receive higher penalties, while nodes with closer neighbors are penalized less.
In addition, this penalty is computed using a Gaussian kernel applied to distances in the feature space that are updated at each propagation step, allowing GRaNDe to adapt to the node representations learned during training.

Experimental results on semi-supervised image classification tasks revealed that replacing standard degree centrality with GRaNDe consistently improved the effectiveness in most evaluated settings, achieving results that are superior or comparable to the state-of-the-art.    As future work, we intend to evaluate GRaNDe in a broader range of scenarios, including citation networks, textual data, and other multimodal settings such as video and audio. We also plan to investigate the integration of GRaNDe with alternative graph construction strategies and propagation mechanisms, as well as its applicability to different GNN architectures beyond diffusion-based models. We also intend to investigate automatic and adaptive strategies for selecting the Gaussian parameter $\sigma$ to better accommodate varying data distributions.

\section*{Acknowledgments}

We thank the São Paulo Research Foundation – FAPESP (grant \#2025/10602-5)  and the University of São Paulo (PRPI Ordinance No. 1032, ``\emph{Apoio aos Novos Docentes}'') for financial support.

\section*{AI Usage Declaration}

The authors used AI-based language models (Claude and Gemini) exclusively for reviewing and improving the clarity of text written by the authors.

\bibliographystyle{sbc}
\bibliography{paper}

\end{document}